\documentclass[runningheads]{llncs}
\usepackage{float}
\usepackage{amsfonts}

\usepackage{amsthm}

\usepackage{amssymb}
\usepackage{amsthm,amsmath}
\usepackage{mathrsfs}
\usepackage{indentfirst}
\usepackage{multirow}
\usepackage{algpseudocode} 
\usepackage[ruled,linesnumbered]{algorithm2e} 

\usepackage{multirow}
\usepackage{times}
\usepackage{soul}
\usepackage{url}
\usepackage[hidelinks]{hyperref}
\usepackage[utf8]{inputenc}
\usepackage[small]{caption}
\usepackage[misc]{ifsym}

\usepackage[T1]{fontenc}
\usepackage{graphicx}
\begin{document}
\title{Resisting Graph Adversarial Attack via Cooperative Homophilous Augmentation}

%

\author{Zhihao Zhu\inst{1} \and
Chenwang Wu\inst{1} \and
Min Zhou\inst{2} \and
Hao Liao\inst{3} \and
DefuLian{\Letter} \inst{1} \and
Enhong Chen\inst{1}}
%
%
\institute{University of Science and Technology of China
\\\email{$\{$zzh98,wcw1996$\}$@mail.ustc.edu.cn
\\$\{$liandefu,cheneh$\}$@ustc.edu.cn }
\and Huawei Technologies co. ltd
\email{zhoum1900@163.com}
\and Shenzhen University
\email{haoliao@szu.edu.cn}
}
%
\toctitle{Resisting Graph Adversarial Attack via Cooperative Homophilous Augmentation}
\tocauthor{}
\maketitle
\begin{abstract}

Recent studies show that Graph Neural Networks(GNNs) are vulnerable and easily fooled by small perturbations, which has raised considerable concerns for adapting GNNs in various safety-critical applications. In this work, we focus on the emerging but critical attack, namely, Graph Injection Attack(GIA), in which the adversary poisons the graph by injecting fake nodes instead of modifying existing structures or node attributes. Inspired by 
findings that the adversarial attacks are related to the increased heterophily on perturbed graphs (the adversary tends to connect dissimilar nodes), we propose a general defense framework CHAGNN against GIA through cooperative homophilous augmentation of graph data and model. 
Specifically, the model generates pseudo-labels for unlabeled nodes in each round of training to reduce heterophilous edges of nodes with distinct labels.  The cleaner graph is fed back to the model, producing more informative pseudo-labels. In such an iterative manner, model robustness is then promisingly enhanced. We present the theoretical analysis of the effect of homophilous augmentation and provide the guarantee of the proposal's validity. 
Experimental results empirically demonstrate the effectiveness of CHAGNN in comparison with recent state-of-the-art defense methods on diverse real-world datasets.

\end{abstract}
%
%
%


\section{Introduction}
In recent years, graph neural networks (GNNs) have been successfully applied in social networks~\cite{zhong2020multiple}, knowledge graphs~\cite{liu2021ragat} and recommender systems~\cite{wu2019session} due to it's good performance in analyzing graph data. 
In spite of the popularity and success of GNNs, they have shown to be vulnerable to adversarial attacks~\cite{li2020adversarial,zhang2020adversarial,ma2020towards}. 
Classification accuracy of GNNs on the target node might be significantly degraded by imperceptible perturbations, posing certain practical difficulties.
For instance, an attacker can disguise his credit rating in credit prediction by establishing links with others.
Financial surveillance enables attackers to conceal the holding relationship in order to carry out a hostile takeover.
Due to the widespread use of graph data, it is critical to design a robust GNN model capable of defending against adversarial attacks.

Existing graph adversarial attacks mainly focus on two aspects \cite{zou2021tdgia}.
The first one is the graph modification attack(GMA), which poisons graphs by modifying the edges and features of original nodes.
The other attack method, graph injection attack (GIA), significantly lowers the performance of graph embedding algorithms on graphs by introducing fake nodes and associated characteristics.
The latter type of attack appears to be more promising.
For instance, it is unquestionably easier for attackers to establish fake users than to manipulate authentic data in recommender systems.
Promoting attackers' influence in social media via registering fake accounts is less likely to be detected than modifying system data.
Given the flexibility and concealment of GIA, it is crucial to develop defense strategies. 
However, the fact is that there are currently fewer methods to defend GIA in comparison to GMA.
In this paper, we propose a defense method against GIA.

Defense methods are mainly categorized into two groups~\cite{akhtar2018threat}.
One approach is to begin with models and then improve their robustness, for example, through adversarial training~\cite{feng2019graph}.
The other seeks to recover the poisoned graph's original data.
Obviously, the first sort of approach is firmly connected to the model, which implies that it may not work with new models.
By contrast, approaches based on data modification are disconnected from concrete models, which is the subject of this paper.

Recent studies \cite{zhu2021relationship,wu2019adversarial} have shown that adversarial attacks are related to the increased heterophily on the perturbed graph, which has inspired works about data cleaning. Existing works \cite{wu2019adversarial} rely heavily on the similarity of features, e.g., utilizing Jaccard similarity or Cosine similarity, to eliminate potentially dirty data. However, they ignored critical local subgraphs in the graph. More specifically, existing methods only measure heterophilous anomalies based on descriptive features while ignoring meaningful interactions between nodes, which leads to biased judgments of heterophily, normal data cleaning, and model performance decline. Additionally, while eliminating heterophily makes empirical sense, it lacks theoretical support.


In this work, we propose a general defense framework, CHAGNN, to resist adversarial attacks by cooperative homophilous augmentation.
To begin, in order to fully use graph information, we propose using GCN labels rather than descriptive attributes to determine heterophily. This is because GCNs' robust representation capability takes into account both the feature and adjacency of nodes. However, it is undeniable that the model must guarantee good performance to provide credible pseudo-labels.
Thus, we further propose a cooperative homophily enhancement of both the model and the graph.
To be more precise, during each round of training, the model assigns prediction labels to the graph data in order to find and clean heterophilous regions, while the cleaned data is supplied back to the model for training in order to get more informative samples.
In this self-enhancing manner, the model's robustness and performance are steadily improved.
Notably, we theoretically demonstrate the effectiveness of homophilous augmentation in resisting adversarial attacks, which has not been demonstrated in prior research~\cite{zhu2021relationship,wu2019adversarial}.
Homophilous augmentation can considerably reduce the risk of graph injection attacks and increase the performance of the model.
The experimental results indicate that after only a few rounds of cleaning, the model outperforms alternative protection approaches.

The contributions of this paper are summarized as follows:
\begin{itemize}
    \item We propose a homophily-augmented model to resist graph injection attacks. 
The model and data increase the graph homophily in a cooperative manner, thereby improving model robustness.
    \item We theoretically prove that the benefit of heterophilous edge removal process is greater than the penalty of misoperation, which guarantees the effectiveness of our method. 
    \item Our experiments consistently demonstrate that our method significantly outperforms over baselines against various GIA methods across different datasets.
\end{itemize}

\section{Preliminaries}
\subsection{Graph Convolutional Network}
Let $G = (V, E)$ be a graph, where $V$ is the set of $N$ nodes, and $E$ is the set of edges.
These edges can be formalized as a sparse adjacency matrix $\boldsymbol{A} \in {\mathbb R}^{N \times N}$, 
and the features of nodes can be represented as a matrix $\boldsymbol{X} \in {\mathbb R}^{N \times D}$, 
where $D$ is the feature dimension.
Besides, in the semi-supervised node classification task, nodes can be divided into labeled nodes $V_L$ and unlabeled nodes $V_U$.

In the node classification task we focus on, the model $f_{\theta}$ is trained based on $G=({\boldsymbol A}, {\boldsymbol X})$ and the labeled nodes $V_L$ to predict all unlabeled nodes $V_U$ as correctly as possible. $ \theta $ is the model's parameter. The model's objective function can be defined as:

\begin{equation}
\max \limits_{\theta} \sum_{v_i\in V_U}^{}I(argmax(f_\theta(G)_i)=y_i),
\end{equation}
where $f_\theta(G)_i \in {[0,1]}^C$, $C$ is the number of categories of nodes.

GCN~\cite{kipf2016semi}, one of the most widely used models in GNNs, aggregates the structural information and attribute information of the graph in the message passing process.
Due to GCN's excellent learning ability and considerable time complexity, it has been applied in various real-world tasks, e.g.,  traffic prediction and recommender systems.
Therefore, it is important to study improve the robustness of GCN against adversarial attacks.
Given $G$ and $V_L$ as input, a two-layer GCN with 
$\theta = (\boldsymbol{W_1}, \boldsymbol{W_2})$ 
implements $f_{\theta} (G)$ as
\begin{equation}
f_\theta(G) = softmax(\boldsymbol{\hat{A}}\sigma(\boldsymbol{\hat{A}}\boldsymbol{X}\boldsymbol{W_1})\boldsymbol{W_2}),
\end{equation}
where $\boldsymbol{\hat{A}} = \boldsymbol{\tilde{D}^{-1/2}(A+I)\tilde{D}^{-1/2}}$ 
and $\boldsymbol{\tilde{D}}$ 
is the diagonal matrix of 
$\boldsymbol{A}+\boldsymbol{I}$.
$\sigma$ represents an activation function, e.g.,  ReLU.

\subsection{Graph Adversarial Attack}
The attacker's goal is to reduce the node classification accuracy of the model on the target nodes $T$ as much as possible.
A poisoning attack on a graph can be formally defined as
\begin{equation}
\min \limits_{G'}\max \limits_{\theta}{\sum_{v_i\in T}^{}I(argmax(f_\theta(G')_i)=y_i)},
\end{equation}
\begin{equation}
s.t. \qquad G'=(\boldsymbol{A',X'}), \left\|\boldsymbol{A'}-\boldsymbol{A}\right\|+\left\|\boldsymbol{X'}-\boldsymbol{X}\right\|\leq\Delta ,\nonumber
\end{equation}
where $\boldsymbol{A'}$ and $\boldsymbol{X'}$ are modified adjacency and feature matrix, and the predefined $\Delta$ is used to ensure that the perturbation on the graph is small enough.

The attacker only makes changes in the original graph without introducing new nodes, which is called graph modification attack (GMA).
Inversely, the attack that does not destroy the original graph but injects new nodes on graphs is defined as the graph injection attack (GIA). 
Modifying existing nodes is often impractical, e.g., manipulating other users in a social network.
However, creating new accounts in social media is feasible and difficult to be detected.
Due to the practicality and concealment of GIA, we focus more on it. 
\begin{figure}[htbp]
        \centering
        \includegraphics[width=0.6\columnwidth]{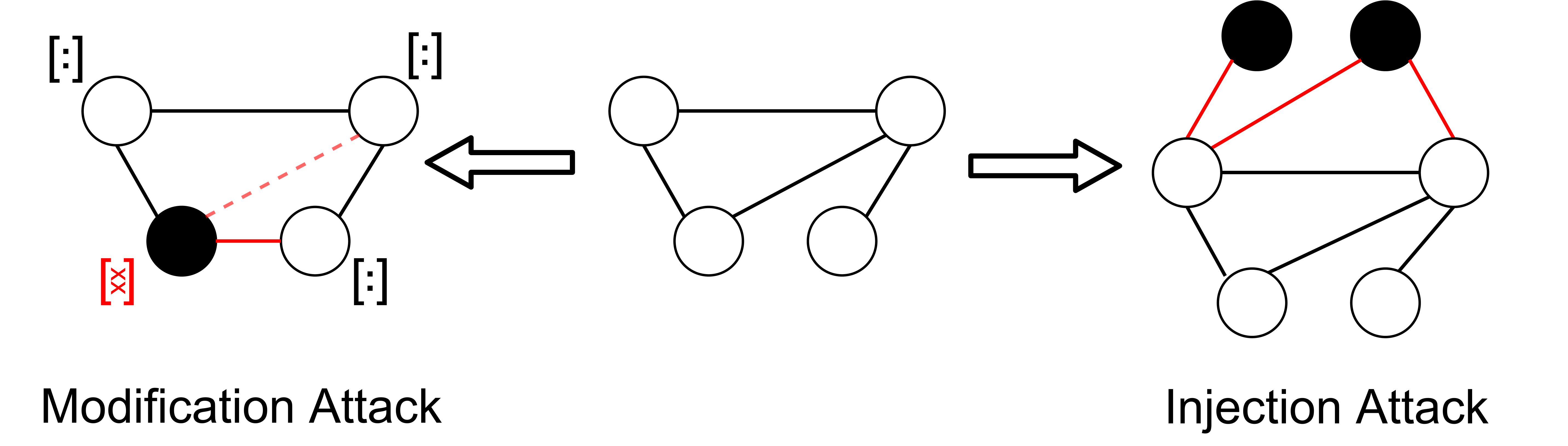}
        \caption{GMA vs GIA}
        \label{gia}
\end{figure}
The difference between GMA and GIA is shown in Fig.\ref{gia}.

Now, we give GIA's formal definition. An attacker is limited to inject $N_I$ nodes with well-crafted features into the graph.
If the injected nodes are represented by $V_I$, then
the injected adjacency matrix and feature matrix can be formalized as follows:

\begin{equation}
\boldsymbol{\mathrm{A}^{\prime}=\left[\begin{array}{cc}
\mathrm{A} & \mathrm{A}_{OI} \\
\mathrm{A}_{OI}^{T} & \mathrm{~A}_{I}
\end{array}\right]}, 
\boldsymbol{\mathrm{A}} \in \mathrm{R}^{N \times N}, \boldsymbol{\mathrm{~A}_{OI}} \in \mathrm{R}^{N \times N_{I}}, \boldsymbol{\mathrm{~A}_{I}} \in \mathrm{R}^{N_{I} \times N_{I}}, \\
\end{equation}

\begin{equation}
\boldsymbol{\mathrm{~X}^{\prime}}=\left[\begin{array}{c}
\boldsymbol{\mathrm{X}} \\
\boldsymbol{\mathrm{X}_{I}}
\end{array}\right], 
\boldsymbol{\mathrm{X}} \in \mathrm{R}^{N \times D}, 
\boldsymbol{\mathrm{~X}_{I}} \in \mathbb{R}^{N_{I} \times D},
\end{equation}
where $\boldsymbol{A_{OI}}$ is the connections between original nodes and injected nodes.

Following the settings of KDD-CUP 2020, $V_U$ and $V_I$ are mixed.
The defender does not know which unlabeled nodes belong to $V_U$ or $V_I$. 
Given $G'$ and $V_L$ as input, the defender's goal is to maximize the classification accuracy of the model on $V_U$.
\begin{equation}
\max \limits_{\theta} \sum_{v_i\in V_U}^{}I(argmax(f_\theta(G')_i)=y_i).
\end{equation}

\section{THE PROPOSED FRAMEWORK}
Based on the practicability and harm of graph injection attacks, we present an efficient method to resist it in this section.
In section 3.1, we relate heterophily to adversarial attacks and defense, and reveal the motivation for our method.
Section 3.2 proposes a defensive framework by homophilous augmentation while leveraging the cooperation of the graph and the model to boost robustness.
Moreover, in section 3.3, we theoretically demonstrate the effectiveness of the proposed method.
 
\subsection{Heterophily and Attack}
Before formally tracing the source of the attack, we give the definition of heterophily and homophily in the graph. If the labels of nodes at both ends of a path are the same, we call it a homophilous path. Conversely, a heterophilous path indicates that the labels of nodes at both ends of it are not the same.
Following~\cite{lim2021new,zhu2020beyond}, we use the homophily ratio $h$ to quantify the degree of homophily, which is defined as the fraction of homophilous edges among all the edges in a graph:
\begin{equation}
h = {|\{(u, v) \in E | y_u = y_v\}|} / {|E|}
\end{equation}
Assume that the nodes in a graph are randomly connected, then for a balanced class, the expectation for $h$ is $\frac{1}{C}$.
If the homophily ratio $h$ satisfies $h>>\frac{1}{C}$, we call the graph a homophilous graph. On the other hand, it is a heterophilous graph if $h<<\frac{1}{C}$. In this paper, we focus on the homophilous graph due to it's ubiquity.

Many research~\cite{wu2019adversarial,jin2020graph} shows that extremely destructive attacks tend to increase the heterophily of the homophilous graph. It seems plausible since neighbor relationships in graph networks provide critical insights for GNN predictions. The attacker cannot destroy these relationships, but can only weaken the connection by connecting heterophilous edges. 
This empirical finding also inspired subsequent research work based on data cleaning. 
We do not have a god-view to know the label of each node, so GCNJaccard~\cite{wu2019adversarial} measures heterophily based on the similarity of features, such as using Jaccard similarity or cosine similarity. Removing heterophilous paths thereby increases the homophily of the graph.
However, they only measure heterophilous anomalies based on descriptive features, ignoring the more critical local subgraphs in the graph. The similarity is stronger if a node and its neighbors share similar hobbies, but it cannot be measured based on descriptive features. Therefore, the unreasonable homophily measures may lead to biased judgments of heterophily and reduce model performance. In addition, these studies are only reasonable assumptions based on experience, and how to guarantee their validity theoretically is challenging.



\begin{algorithm}[t]
\caption{Eliminating heterophilous Edges}  
\LinesNumbered  
\KwIn{Poisoned graph $G'$, modified nodes $V_M$,  labeled nodes $V_L$, pseudo-labels $\hat{Y}$, soft-labels $f_{\theta}(G')$, elimination rate $q$}
\KwOut{Cleaned Graph $\hat{G'}$}
$H_e = \emptyset$\\
\For{$u \in V_M$}
{
	$\hat{y_u}\xleftarrow[]{}$ pseudo-label of node $u$;
	$N_u\xleftarrow[]{}$ $u$'s neighbors\\
	\For{$v \in N_u$}
	{
    \If {$(v \in V_L$ and $y_v \neq \hat{y_u})$ \textbf{or} $(v \notin V_L$ and $\hat{y_v} \neq \hat{y_u})$}
    {
		$H_e \xleftarrow[]{} H_e \cup \{(u,v)\}$
	}
	}
}
\For{$(u,v) \in H_e$}
{
    $f_{\theta}(G')_{u} \xleftarrow[]{}$ soft label of node $u$;
    $f_{\theta}(G')_{v} \xleftarrow[]{}$ soft label of node $v$\\
	The degree of heterophily of $(u,v)$ is 
	$\bar{h}_{u,v} = JS(f_{\theta}(G')_{u},f_{\theta}(G')_{v})$
}
Pick and eliminate $q \cdot |H_e|$ heterophilous edges according to the sampling probability vector $\boldsymbol{p}$, which is calculated by Eq.(9)\\
Output the cleaned graph $\hat{G'}$
\end{algorithm} 

\subsection{Cooperative homophilous Augmentation}
Considering the deficiencies of existing methods discussed in last subsection, we propose a synergistic homophily augmentation strategy to resist attacks.
As mentioned before, using the similarity of features without graph's structure information to represent heterophily is biased. 
Thus, we propose to increase the graph’s homophily by pseudo-labels that contain the information of both features and structure.

In GIA scenarios, fake connections must be the edges of unlabeled nodes.
Therefore, our method focuses on this region of the graph.
Due to the gap between pseudo-labels and labels, using pseudo-labels to discriminate and remove heterophilous edges may lead to mistakenly eliminating homophilous edges. 
Besides, pseudo-labels can not quantify the strength of heterophily of edges.
For example, suppose the predictions of nodes $u$, $v$, and $w$ are $[0.99, 0.01]$, $[0.49, 0.51]$ and $[0.01,0.99]$ respectively.
Pseudo-labels will treat edge ($u$,$v$) and ($u$,$w$) as identical, which is unreasonable.
Compared to pseudo-labels, a node's soft label can more specifically reflect the probability that the node belongs to each category.
Therefore, we use the $JS$ divergence of the soft labels of nodes at both ends of the edge to measure the degree of heterophily of the edge ($u$,$v$).
Heterophilous edges with a high degree of heterophily are more likely to be removed.

\begin{align}
\bar{h}_{u,v}=&JS\left(f_{\theta}(G')_{u}, f_{\theta}(G')_{v}\right) \nonumber\\
=&\frac{1}{2} \sum_{i=1}^{C} f_{\theta}(G')_{u}^{i} \log \frac{2f_{\theta}(G')_{u}^{i}}{f_{\theta}(G')_{u}^{i}+f_{\theta}(G')_{v}^{i}}
+\frac{1}{2} \sum_{i=1}^{C} f_{\theta}(G')_{v}^{i} \log \frac{2f_{\theta}(G')_{v}^{i}}{f_{\theta}(G')_{u}^{i}+f_{\theta}(G')_{v}^{i}} ,
\end{align}
where $f_{\theta}(G')_{u}$ is the soft label of node $u$,
$f_{\theta}(G')_{u}^{i}$ denotes the probability that node $u$ belongs to class $i$.

The value range of $JS$ divergence is $[0,1]$.
The value of the $JS$ divergence is closer to 0 as the two probability distributions are more similar.
It means that the smaller the value, the more likely the edge is a homophilous edge.
We normalize the vector $\bar h$, which stores the degree of heterophily of all the heterogeneous edges.
\begin{equation}
\boldsymbol{p_{i,j}} = exp({\bar h_{i,j}}) / \sum_{(u,v) \in H_e} exp({\bar h_{u,v}}),
\end{equation}
where $H_e$ is the heterophilous edges set, $p_{i,j}$ is the probability that $(i,j)$ is sampled to be removed.
Then we pick out some heterophilous edges according to the sampling probability vector $\boldsymbol{p}$.
Edges with a higher degree of heterophily are more likely to be picked out for removal.
The process of eliminating heterophilous edges is described in Algorithm 1.

However, the result of Algorithm 1 strongly depends on the authenticity of the pseudo-label.
To achieve better performance, we propose to enhance the homophily of graph via cooperatively cleaning graph and improving model performance.
Specifically, the model provides pseudo-labels to clean the data,
while the purified graph guides the model by providing more reliable pseudo-labels.
The model and data thus cooperatively increase classification accuracy.

\begin{figure}[htbp]
        \centering
        \includegraphics[width=\columnwidth]{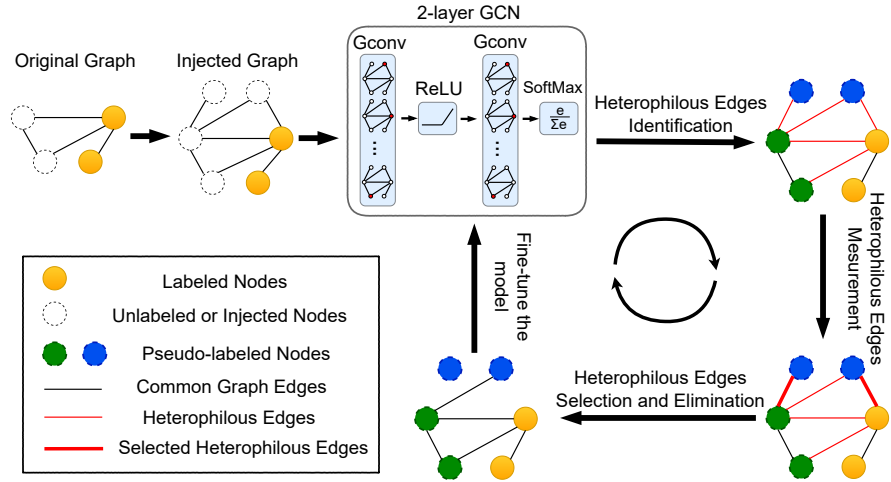}
        \caption{Algorithm Flowchart of CHAGNN}
        \label{alg}
\end{figure}

Next we give the implementation details of CHAGNN. 
In GIA scenarios, the poisoned regions are consumingly related to the unlabeled nodes, including the unlabeled nodes $V_U$ in the original graph and injected nodes $V_I$.
The nodes selected to modify their edges are called modified nodes ($V_M$).
In order to accurately remove maliciously injected edges, we simply define $V_M$ as $V_U \cup V_I$ .
At the beginning of our algorithm, we first use poisoned graph to conduct pre-training process on the model.
Then we obtain all nodes' pseudo-labels and soft labels.
The pseudo-labels and soft labels are input to Algorithm 1 to generate a purified graph $\hat{G}$.
After that, model parameters will be fine-tuned on $\hat{G}$.
This process will dramatically improve classification performance in a few rounds.
The details of our method are shown in Algorithm 2.
The algorithm flowchart of CHAGNN is shown in Fig.\ref{alg}.
\begin{algorithm}[t]
\caption{CHAGNN}  
\LinesNumbered  
\KwIn{Poisoned graph $G'$, labeled nodes $V_L$, modified nodes $V_M$, elimination rate $q$, max iterations $max\_iter$}
\KwOut{Prediction on test set}
Pretrain model parameters $\theta$ on $G'$\\
\For{i={1,...,max\_iter}}
{
	Obtain the pseudo-labels $P$ and soft-labels $f_{\theta}(G')$ of all nodes\\
    $G' \xleftarrow[]{} Algorithm 1(G', V_M, V_L, P, f_{\theta}(G'), q)$\\
	Fine-tune $\theta$ on $G'$\\
}
    Output the prediction on test set
\end{algorithm} 

We found that AdaEdge~\cite{chen2020measuring} also used pseudo-labels in their algorithm, but our proposal is quite different from it. 
Unlike AdaEdge, which directly removes the heterophilous edges based on pseudo labels, we introduce $JS$ divergence to quantify the degree of heterophily of heterophilous edges before the elimination process, which greatly reduces the possibility of misoperation. 
Besides, the subjects of the research are different.
AdaEdge focuses on solving the over-smoothing problem, and they perform a cleanup operation on the entire graph.
Instead, we consider the scenario of graph injection attacks, applying heterophilous edge removal to potentially injected edges in the graph and providing the corresponding theoretical guarantee.
The experimental comparison results of the two algorithms are shown in section 4.2 and 4.3.

\subsection{Theoretical Guarantee}
In general, removing heterophilous edges benefits model, while homophilous edges deletion brings model penalties.
We want to mitigate the damage of the graph data by heterophilous edges as much as possible.
However, identifying and eliminating heterophilous edges via nodes' pseudo-labels may mistakenly delete homophilous edges.
In this section, we prove that given an arbitrary model accuracy, the expected benefits of the proposed strategy outweigh the expected penalties.
Specifically, we firstly use the variation of loss to represent the impact of eliminating heterophilous(homophilous) edges.
Then we analyze the probability of deleting homophilous edge by mistake with model's accuracy.
Combining these two parts, we guarantee the reliability of CHAGNN theoretically. We give the proof of theorems in appendix.

To simplify the proof, we employ the SGC model which removes the activation function compared with GCN model.
Given $G = (\boldsymbol{A, X})$ and $Y_L$ as input, a two-layer SGC with $\theta = \boldsymbol{W}$ implements $f_\theta(G)$ as
\begin{equation}
f_\theta(G) = softmax(\boldsymbol{\hat{A}^2XW}).
\end{equation}
Following ~\cite{zhu2021relationship}'s setting, we assume that $G$ is a d-regular graph which means that each node of $G$ has $d$ connections
with other nodes.
For each node of $G$, proportion $h$ of their neighbors belong to the same class, 
while proportion $\frac{1 - h}{C - 1}$ of them belong to any other class uniformly.
The features of node $v$ are defined as  $\boldsymbol{x_v} = p \cdot onehot(\boldsymbol{y_v}) + \frac{1 - p}{ C }$, where $\boldsymbol{y_v}$ means the node's label.

We use the change of CM loss of the model to analyze the influence of injecting nodes to the graph.
The CM loss of node $v$ is defined as:
\begin{equation}
\boldsymbol{Z=\hat{A}^2XW}, loss_{v} = \boldsymbol{Z}_{vy_{v}} - \max \limits_{j\neq y_{v}} \boldsymbol{Z}_{vj}.
\end{equation}
Define the CM loss of node $v$ on clean graph as $L_0$.
After we generate  nodes to inject homophilous edges to the graph, the CM loss changes to $L_1$.
Correspondingly, the CM loss is called $L_2$ after we generate nodes to inject heterophilous edges to the graph.
Assume that the proportion of node $v$'s edges which is connected to class $y_v$ before poisoned is $h_0$(including self-loop of node $v$), and the proportion of other classes is $h_1$.
After we inject nodes to the graph, the proportion of node $v$'s edges which is connected to class $y_v$ is $r_0$, and the proportion of other classes is $r_1$.
For convenience, we separate the proportions of injected edges from $r_0$ or $r_1$. 
The proportion of injected edges is denoted as $r_2$. 
\begin{theorem}
Consider target attack and direct attack which means that the inject nodes are directly connected to the target node $v$.
Then we have:
\begin{equation}
\frac{L_1-L_0}{L_0-L_2} = \frac{(r_0 - r_1 + r_2) - (h_0 - h_1)}{(h_0 - h_1) - (r_0 - r_1 - r_2)}
\end{equation}
\end{theorem}
\begin{remark}
heterophilous edges elimination is actually the reverse process of attack. 
According to Theorem 1, we can estimate the ratio between the penalty of deleting a homophilous edge and the benefit of deleting a heterophilous edge. 
\end{remark}

Based on the relation between the penalty  and benefit stated in Theorem 1, 
we analyze the expected benefit and expected penalty of the edge deletion operation at the specified model accuracy.
For simplicity, we focus on judging whether a node belongs to a specific class in theorem 2, which is a binary classification problem.
Referring to~\cite{mohri2018foundations}, it is easily extensible and applicable to multi-class scenarios.

Assume that the prediction accuracy of model on unlabeled nodes is $p$. 
The prediction accuracy of different nodes is independent.
Suppose we judge that there is a heterogeneous edge between nodes $u$ and $v$ according to pseudo-labels and then we delete $e_{uv}$.
The probability that $e_{uv}$ is actually a homogeneous edge is $p_1$ 
while the probability that $e_{uv}$ is actually a heterogeneous edge is $p_2$.
The pseudo-labels of $u$ and $v$ are $\hat{y_u}$ and $\hat{y_v}$, $\hat{y_u} \neq \hat{y_v}$.
The labels of $u$ and $v$ are $y_u$ and $y_v$.
Then we have:

\begin{theorem}
The ratio of expected penalty to expected benefit for eliminating an edge in CHAGNN is related to the prediction accuracy $p$.
\begin{align}
\frac{e_1}{e_2} < 2p(1-p) < 1 
\end{align}
\end{theorem}
\begin{remark}
Theorem 2 shows that the expected benefit is always greater than the expected penalty in our algorithm.
For a binary classification problem, an effective classifier should have an accuracy greater than 50\%. 
This means that we can reduce the ratio in Theorem 2 by continuously improving the accuracy of an effective classifier.
\end{remark}

\section{Experiment}
In this section, we compare the proposed CHAGNN with state-of-the-art defense strategies.
The experiment primarily validates our algorithm’s excellent performance by answering the following research questions:

\begin{itemize}
    \item RQ1.How well does CHAGNN perform compared to other state-of-the-art defense methods under different graph injection attacks?
    \item RQ2.How well does CHAGNN perform with different injected nodes ratio under the state-of-the-art GIA methods?
    \item RQ3.How much does the deletion rate affect the performance of CHAGNN?
\end{itemize}


\subsection{Experimental setup}
\subsubsection{Dataset}
We evaluate the proposed algorithms with four widely used citation network datasets, including Cora-ml, Cora~\cite{bojchevski2017deep,mccallum2000automating}, Citeseer~\cite{giles1998citeseer}, and Pubmed.
The statistics of datasets are summarized in Table~\ref{Data}. Following~\cite{zugner2019adversarial}, we only consider the largest connected component (LCC) of each graph data. 

To evaluate the effectiveness of our method, we compared it with the state-of-the-art defense models.
The compared algorithms and attack methods are introduced in the next two subsections.
\begin{table}[htbp]
\centering
\caption{Statistics of benchmark datasets}
\label{Data}
\renewcommand{\arraystretch}{1.1}
\resizebox{0.6\textwidth}{!}{
\setlength{\tabcolsep}{0.02\linewidth}{
\begin{tabular}{cclll}
\hline
         & $N_{LCC}$  & $E_{LCC}$  & Classes & Features \\
\hline
Cora-ml  & 2810  & 7981  & 7       & 2879     \\
Cora     & 2485  & 5069  & 7       & 1433     \\
Citeseer & 2110  & 3668  & 6       & 3703     \\
Pubmed   & 19717 & 44338 & 3       & 500      \\
\hline
\end{tabular}}}
\end{table}

\subsubsection{Compared Algorithms}
\begin{itemize}
    \item GCN~\cite{kipf2016semi}: We compare our algorithm with other methods with GCN, one of the most widely used models in GNNs.
    \item GCNSVD~\cite{entezari2020all}: GCNSVD is a preprocessing method to resist adversarial attacks. It use a low-rank approximation of the graph to train GCN.
    \item GCN-Jaccard~\cite{wu2019adversarial}: Another preprocessing method to resist adversarial attacks. They identity and eliminate heterophilous edges with nodes' features.
    \item GNNGUARD~\cite{zhang2020gnnguard}: GNNGUARD added the attention mechanism to defend against adversarial attacks.
It learns how to best assign higher weights to edges connecting similar nodes while pruning edges between unrelated nodes.
    \item ORH~\cite{zhu2021relationship}: ORH mitigates the damage to the graph structure on account of the addition of heterophilouss edges by increasing the node's weight.
    \item VPN~\cite{jin2021power}: VPN replaces the graph convolutional operator $\boldsymbol A$ with the weighted sum of adjacency matrices with different powers. 
    \item AdaEdge~\cite{chen2020measuring}: AdaEdge uses pseudo-labels to remove heterophilous edges to solve model's over-smoothing problem. Unlike our method, AdaEgde does not consider actual attack scenarios. Moreover, the judged heterophilous edges are directly removed without screening, which can easily lead to the mistaken deletion of homophilous edges.
\end{itemize}

\subsubsection{Attack Methods}
\begin{itemize}
    \item TDGIA~\cite{zou2021tdgia}: TDGIA first introduces the topological defective edge selection strategy to choose the original nodes for connecting with the injected ones. It then designs the smooth feature optimization objective to generate the features for the injected nodes.
    \item FGA~\cite{chen2018fast}: A framework to generate adversarial networks based on the gradient information in GCN.
    \item MGA~\cite{chen2020mga}: 
This paper proposes a Momentum Gradient Attack (MGA) against the GCN model, which can achieve more aggressive attacks with fewer rewiring links than FGA.
\end{itemize}
FGA and MGA are not directly applicable in GIA scenario.
We modify them to work for GIA setting. 
They are performed on the graph poisoned by a heuristic injection.

\begin{table}[htbp]
\centering
\caption{Node classification performance (Accuracy $\pm$ Std) under non-targeted attack}\label{Tabel 1}
\renewcommand{\arraystretch}{1.}
\setlength{\tabcolsep}{0.015\linewidth}{
\begin{tabular}{llllll}
\hline\noalign{\smallskip}
\small
Attack & Defense & Cora-ml    & Cora       & Citeseer   & Pubmed     \\ \noalign{\smallskip}\hline\noalign{\smallskip}
TDGIA          & No Attack       & 84.64$\pm$0.34 & 81.26$\pm$1.20 & 71.46$\pm$0.30 & 85.00$\pm$0.09  \\
               & Attack          & 67.80$\pm$0.44 & 71.52$\pm$0.33 & 60.18$\pm$1.19 & 73.18$\pm$0.25 \\
               & GCNSVD          & 68.44$\pm$0.28 & 73.56$\pm$0.26 & 63.24$\pm$0.55 & 78.06$\pm$0.28 \\
               & GCNJaccard      & 65.10$\pm$1.35 & 70.04$\pm$0.48 & 61.60$\pm$0.70 & 76.48$\pm$0.23 \\
               & GNNGUARD        & 65.26$\pm$1.35 & 72.30$\pm$0.23 & 64.38$\pm$0.61 & 70.46$\pm$0.66 \\
               & ORH             & 56.30$\pm$1.13 & 65.16$\pm$0.55 & 55.92$\pm$1.34 & 70.58$\pm$0.46 \\
               & VPN             & 69.03$\pm$1.11 & 75.76$\pm$0.39 & 66.90$\pm$0.83 & 78.83$\pm$0.17 \\
               & AdaEdge         & 76.34$\pm$1.43 & 76.86$\pm$0.48 & 67.62$\pm$0.44 & 78.62$\pm$0.20  \\
               & \textbf{CHAGNN}          & \textbf{79.52$\pm$0.32} & \textbf{77.84$\pm$0.29} & \textbf{69.22$\pm$0.59} & \textbf{79.66$\pm$0.37} \\ \noalign{\smallskip}\hline\noalign{\smallskip}
FGA            & No Attack       & 84.64$\pm$0.34 & 81.26$\pm$1.20 & 71.46$\pm$0.3  & 85.00$\pm$0.09  \\
               & Attack          & 82.08$\pm$0.33 & 79.38$\pm$0.58 & 70.72$\pm$0.56 & 80.62$\pm$0.32 \\
               & GCNSVD          & 78.84$\pm$0.16 & 76.66$\pm$0.22 & 67.66$\pm$0.3  & 80.96$\pm$0.05 \\
               & GCNJaccard      & 79.66$\pm$0.91 & 79.38$\pm$0.32 & \textbf{70.92$\pm$0.27} & 81.52$\pm$0.17 \\
               & GNNGUARD        & 74.98$\pm$0.41 & 74.42$\pm$0.55 & 69.26$\pm$0.84 & 80.38$\pm$0.07 \\
               & ORH             & 73.34$\pm$1.61 & 74.22$\pm$0.84 & 66.70$\pm$2.74 & 72.20$\pm$0.25  \\
               & VPN             & 78.53$\pm$0.71 & 74.32$\pm$0.64 & 69.76$\pm$0.62 & 81.48$\pm$0.13 \\
               & AdaEdge         & 83.00$\pm$0.41 & 78.66$\pm$0.66 & 70.08$\pm$1.12 & 81.40$\pm$0.14  \\
               & \textbf{CHAGNN}          & \textbf{83.06$\pm$0.56} & \textbf{79.64$\pm$0.6}  & 70.36$\pm$0.83 & \textbf{81.56$\pm$0.14} \\ \noalign{\smallskip}\hline\noalign{\smallskip}
MGA            & No Attack       & 84.64$\pm$0.34 & 81.26$\pm$1.20 & 71.46$\pm$0.30 & 85.00$\pm$0.09  \\
               & Attack          & 81.98$\pm$0.70 & 76.06$\pm$0.54 & 69.68$\pm$0.45 & 80.74$\pm$0.10  \\
               & GCNSVD          & 80.38$\pm$0.30 & 74.42$\pm$0.47 & 67.82$\pm$0.17 & 80.96$\pm$0.05 \\
               & GCNJaccard      & 80.10$\pm$0.82 & 75.42$\pm$0.43 & 69.94$\pm$0.38 & 81.30$\pm$0.44  \\
               & GNNGUARD        & 73.56$\pm$0.37 & 72.18$\pm$0.55 & 67.28$\pm$0.77 & 80.62$\pm$0.69 \\
               & ORH             & 72.46$\pm$1.66 & 69.24$\pm$0.95 & 66.68$\pm$1.20 & 74.66$\pm$0.48 \\
               & VPN             & 78.93$\pm$0.84 & 74.68$\pm$0.36 & 69.84$\pm$0.64 & 81.20$\pm$0.05 \\
               & AdaEdge         & 83.26$\pm$0.67 & 77.64$\pm$0.68 & 70.04$\pm$0.27 & 81.12$\pm$0.12 \\
               & \textbf{CHAGNN}          & \textbf{83.66$\pm$0.62} & \textbf{78.02$\pm$0.23} & \textbf{70.14$\pm$0.34} & \textbf{81.38$\pm$0.15} \\ \hline
\end{tabular}}
\end{table}

\subsubsection{Parameter Settings}
For each dataset, we randomly split the nodes into labeled nodes for training procedure(10\%),
labeled nodes for validation(10\%) , and unlabeled nodes as test set to evaluate the model(80\%).
The hyper-parameters of all the models are tuned based on the loss and accuracy on validation set.
We report the average performance of 5 runs for each experiment.
To avoid excessive cleaning of the graph, we fixed the elimination rate in each iteration at 10\% and the maximum number of iterations at 5.

\subsection{Defense Performance Against Non-targeted Adversarial Attacks}
We compare the performance of different methods at 10\% injected nodes rate on four datasets.
The results are shown in Table~\ref{Tabel 1}. 
We highlight the best performance in bold. 
From the table, we have the following observations and discussions.
\begin{itemize}
    \item CHAGNN significantly outperforms all compared algorithms on most settings, indicating that the validity of targeted design towards GIA and cooperative homophilous augmentation.
    \item The performance of FGA and MGA is not significant compared to TDGIA.
It makes sense because TDGIA was designed specifically for GIA scenarios, whereas FGA and MGA were originally designed for GMA. 
When resisting weak attacks, such as FGA and MGA, the defense performance of several compared models is poor or even worse than the vanilla GCN.
We think it is due to the fact that the graph considered by the defense algorithm is severely damaged.
However, when dealing with less poisoned or clean graphs, the performance of most defense algorithms may decrease.
For instance, GCNSVD uses low-rank representation of the graph, leading to the loss of information carried in the original graph structure. 
The performance of GCNSVD on the original graph will be worse than the vanilla GCN.
We can also see this phenomenon in ~\cite{jin2020graph}'s experiment.
    \item The performance of ORH and VPN fluctuates greatly. We think it is because the performance of both algorithms depends on the choice of hyperparameters. 
    Specifically, the performance of ORH depends on the weight of the node's own information and neighbors' information in the message passing process. 
    The performance of VPN depends on the weights of different powered graphs.
\end{itemize}

\subsection{Defense Performance Under Different Injected Nodes Ratio}
\begin{figure}[htbp]
        \centering
        \includegraphics[width=0.8\columnwidth]{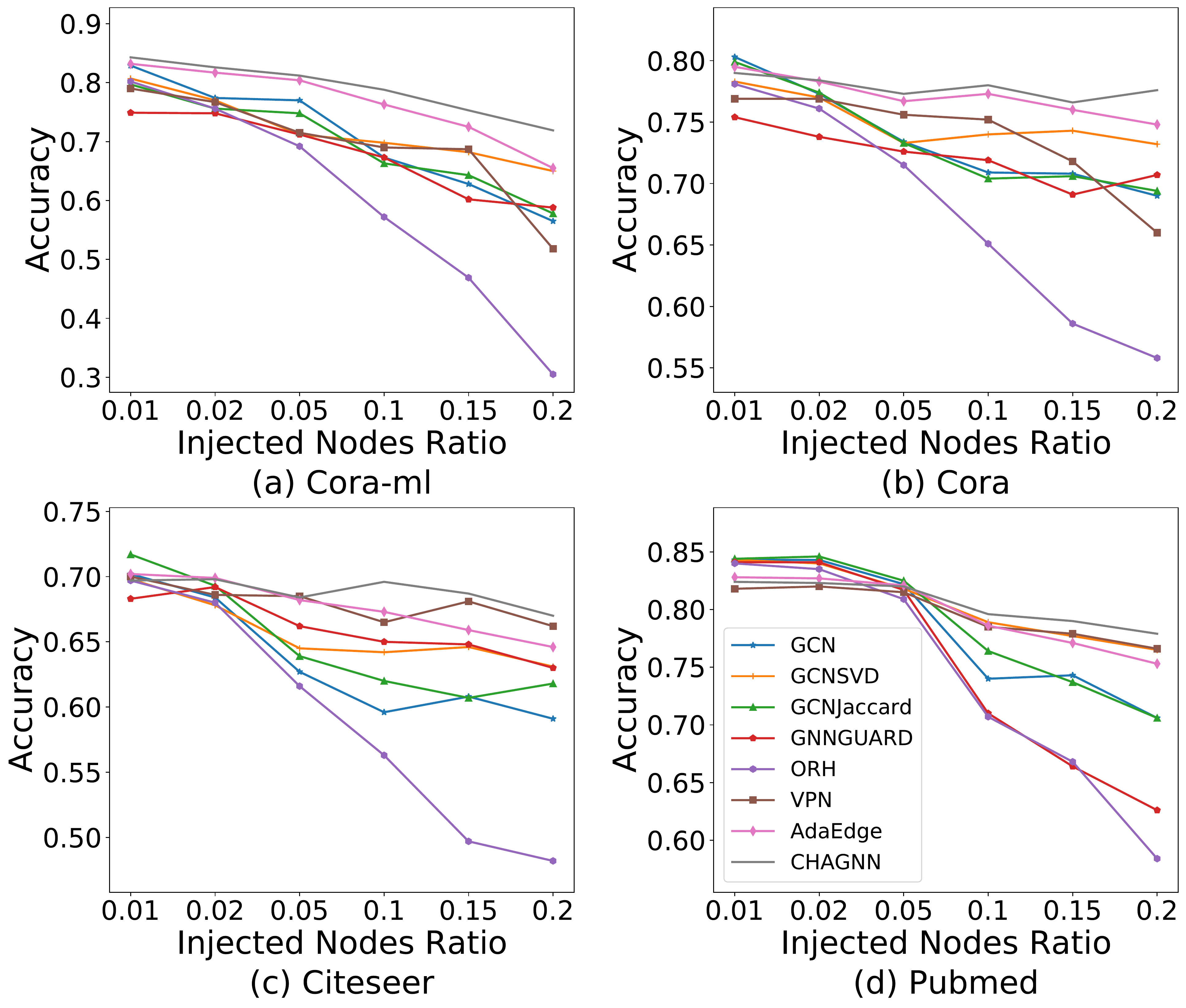}
        \caption{Node classification performance under different injected nodes ratio on Cora-ml, Cora, Citeseer and Pubmed}
        \label{inject_rate3}
\end{figure}

We compare the performance of different algorithms under different injected nodes ratio.
We choose TDGIA, the attack method with the best results in our experiment,
to evaluate the performance of defense methods under different injected nodes ratios.
The results are reported in Fig.\ref{inject_rate3}.
Observations and discussions are listed as follows.
\begin{itemize}
    \item Our method is effective against more powerful attacks.
    Even with a high injected nodes ratio, our approach can significantly improve model's performance.
Vanilla GCN shows poor performance under 20\% injected nodes ratio.
Our method can improve it by 27\%, 12\%, 13\% and 10\% on the four datasets respectively. 
    \item TDGIA shows better performance as the injected nodes ratio increases.
Under different injected nodes ratios, our method outperforms others in most cases, exhibiting excellent defensive performance. 
It illustrates that heterophilous edges elimination can indeed enhance the robustness of the model against adversarial attacks.
    \item The performance of AdaEdge is better than GCNJaccard, which illustrates the effectiveness of using nodes' pseudo-labels to discriminate heterophily is more effective than using nodes' features.
    The performance of AdaEdge is second only to CHAGNN on multiple datasets. It shows that the process of screening the discriminated heterophilous edges can effectively reduce the possibility of homophilous edges being mistakenly removed, which brings stronger defense performance in CHAGNN.
\end{itemize}

\subsection{Parameter Sensitivity on Eliminating Rate}
\begin{figure}[htbp]
        \centering
        \includegraphics[width=0.8\columnwidth]{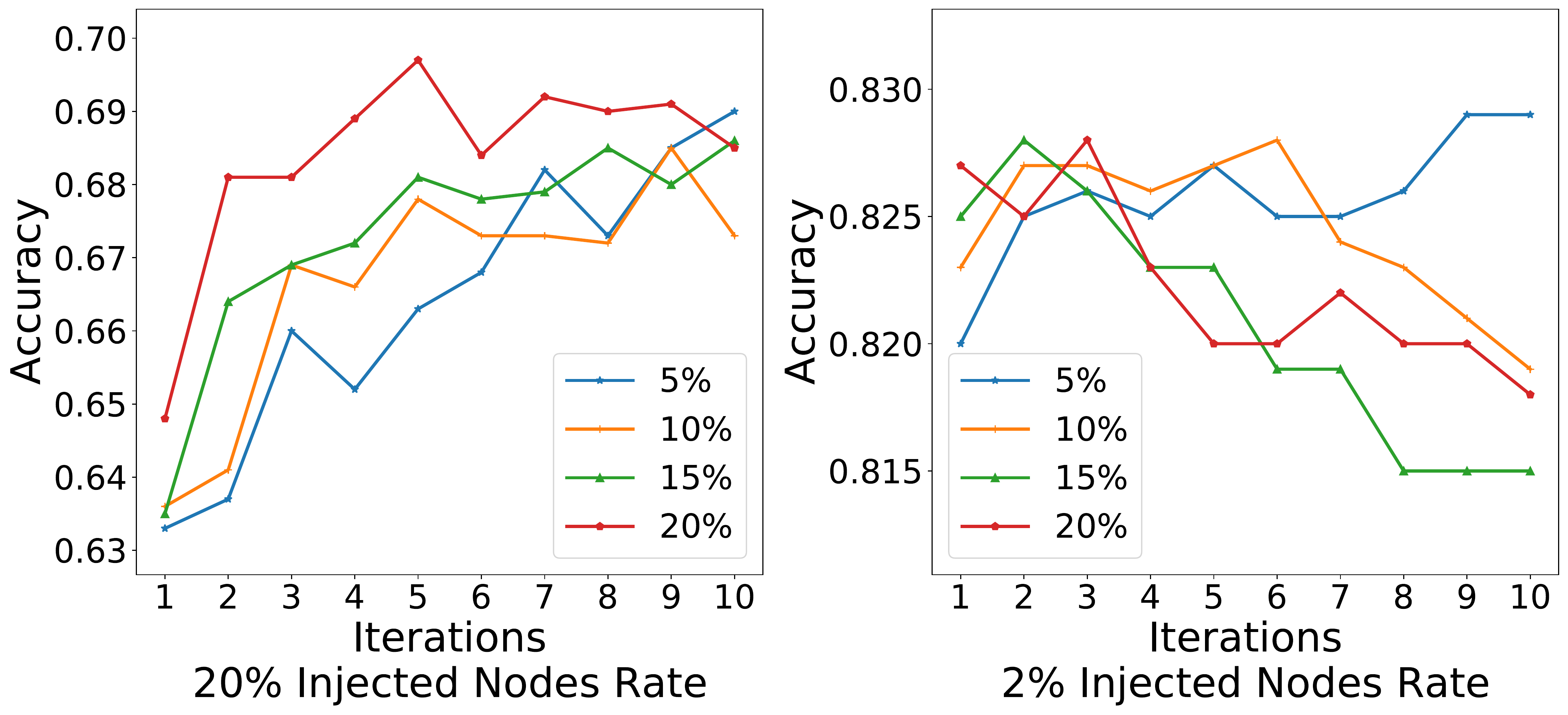}
        \caption{Node classification performance under different eliminating rates}
        \label{dele}
\end{figure}

In this part, we conduct sensitivity analysis with respect to the eliminating rate. 
We only report the results for the Cora-ml dataset at 20\% and 2\% injected nodes rates, since the results for other datasets share similar trends. 
The performance of node classification with different eliminating rates under TDGIA is shown in Fig.\ref{dele}.
We fixed the maximum number of iterations at 10.
The following are some observations.
\begin{itemize}
    \item The classification performance improves overall as the number of iterations increases.
Our method has a certain defensive effect on most eliminating rates.
    \item It is not true that the higher the eliminating rate, the better our method performs.
An excessive eliminating rate on a graph with few injected nodes can cause our method to perform poorly.
This is because in a graph with few injected nodes, it is very easy to remove homophilous edges by mistake.
In the future, we will devise some efficient methods to find the most appropriate eliminating rate.
\end{itemize}

\section{Related Work}


\subsection{Adversarial Attacks on GNNs}
Nettack~\cite{zugner2018adversarial} stated that adding unnoticeable perturbations to the graph can fool GCN into incorrectly predicting.
They generated perturbations to lead GCN to misclassify the target node while preserving the features' co-occurrences and the graph's degree distribution.
Metattack~\cite{zugner2019adversarial} is proposed to reduce the overall performance of the model based on meta-learning.
Most attacks are based on modifying nodes in the original graph.
A more realistic scenario, graph injection attack (GIA), is studied in~\cite{wang2020scalable,sun2020adversarial}, which injects new vicious nodes instead of modifying the original graph.
A greedy algorithm~\cite{wang2018attack} is proposed to generate edges of malicious nodes and their corresponding features aiming to minimize the classification accuracy on the target nodes.
NIPA~\cite{sun2020adversarial} modeled the critical steps of graph injection attack based on reinforcement learning strategy.
TDGIA~\cite{zou2021tdgia} presented an analysis on the topological vulnerability of GNNs under GIA setting and proposed the topological defective graph injection attack (TDGIA) for effective injection attacks.

\subsection{Defenses on GNNs}
GCNSVD~\cite{entezari2020all} found that Nettack has a greater impact on the high-rank part of the network.
Then they proposed to use a low-rank approximation of the graph to train GCN by Singular Value Decomposition(SVD).
GCNJaccard~\cite{wu2019adversarial} stated that the attacks tend to connect the target node to nodes with different features. 
They removed the edges connecting the nodes that share few similarities to the target node by jaccard similarity.
Pro-GNN~\cite{jin2020graph} explored both properties mentioned before and designed a general framework to jointly learn a structural graph and a robust graph neural network model guided by these properties.
GNNGUARD~\cite{zhang2020gnnguard} detected and quantified the relationship between the graph structure and node features and then exploited that relationship to mitigate negative effects of the adversarial attacks.
In addition to the defense methods against graph adversarial attacks , some methods based on data augmentation can also mitigate the influence of the model on graph adversarial attacks.
VPN~\cite{jin2021power} designed the robust GCN via graph powering. They proposed a new convolution operator that is provably robust in the spectral domain.
They incorporated it in the GCN architecture to improve model's expressivity and interpretability.
AdaEdge~\cite{chen2020measuring} optimizes the graph topology based on the model predictions for relieving the over-smoothing issue. They simply remove the heterophilous edges without considering the effect of mistakenly removing the homophilous edges in this process.
And the method does not consider the scenario of graph adversarial attacks.
GAUG~\cite{zhao2020data} used GAE to help improve GCN's robustness.The model's effectiveness at defending graph adversarial attacks depends on GAE's performance.
However, all the defense methods mentioned are designed for GMA.
As there are currently few methods to defend GIA, this paper defines this problem, which may provide critical insights for future research.

\section{Conclusion}
A more realistic scenario, graph injection attack (GIA), demonstrated effective attack performance on GNNs.
However, there were few specific defense methods against GIA, a scenario that is easier for attackers to implement.
In this paper, we formalized the anti-GIA defense scenario and designed the corresponding algorithm.
Our experiments showed that our method significantly outperforms state-of-the-art baselines and improves the overall robustness under various GIA methods. 
Theoretically, the proposed strategy could work in various graph adversarial attacks. However, in the more practical GIA scenario, we can strictly guarantee the effectiveness from empirical and theoretical aspects. In the future, we plan to apply this strategy to more attack scenarios.




\begin{thebibliography}{8}
\bibitem{zhong2020multiple}
Zhong T, Wang T, Wang J, et al. Multiple-aspect attentional graph neural networks for online social network user localization[J]. IEEE Access, 2020, 8: 95223-95234.

\bibitem{liu2021ragat}
Liu X, Tan H, Chen Q, et al. RAGAT: Relation aware graph attention network for knowledge graph completion[J]. IEEE Access, 2021, 9: 20840-20849.

\bibitem{wu2019session}
Wu S, Tang Y, Zhu Y, et al. Session-based recommendation with graph neural networks[C]//Proceedings of the AAAI'19, 33(01): 346-353.

\bibitem{li2020adversarial}
Li J, Zhang H, Han Z, et al. Adversarial attack on community detection by hiding individuals[C]//Proceedings of The Web Conference 2020. 2020: 917-927.

\bibitem{zhang2020adversarial}
Zhang M, Hu L, Shi C, et al. Adversarial label-flipping attack and defense for graph neural networks[C]//Proceedings of ICDM'20. IEEE, 791-800.

\bibitem{ma2020towards}
Ma J, Ding S, Mei Q. Towards more practical adversarial attacks on graph neural networks[J]. Advances in neural information processing systems, 2020, 33: 4756-4766.

\bibitem{zou2021tdgia}
Zou X, Zheng Q, Dong Y, et al. TDGIA: Effective injection attacks on graph neural networks[C]//Proceedings of KDD'21. 2461-2471.

\bibitem{akhtar2018threat}
Akhtar N, Mian A. Threat of adversarial attacks on deep learning in computer vision: A survey[J]. Ieee Access, 2018, 6: 14410-14430.

\bibitem{feng2019graph}
Feng F, He X, Tang J, et al. Graph adversarial training: Dynamically regularizing based on graph structure[J]. IEEE Transactions on Knowledge and Data Engineering, 2019, 33(6): 2493-2504.

\bibitem{zhu2021relationship}
Zhu J, Jin J, Loveland D, et al. On the Relationship between Heterophily and Robustness of Graph Neural Networks[J]. arXiv preprint arXiv:2106.07767, 2021.

\bibitem{zugner2018adversarial}
Zügner D, Akbarnejad A, Günnemann S. Adversarial attacks on neural networks for graph data[C]//Proceedings of KDD'18. 2847-2856.

\bibitem{zugner2019adversarial}
Zügner D, Günnemann S. Adversarial attacks on graph neural networks via meta learning[J]. arXiv preprint arXiv:1902.08412, 2019.

\bibitem{sun2020adversarial}
Sun Y, Wang S, Tang X, et al. Adversarial attacks on graph neural networks via node injections: A hierarchical reinforcement learning approach[C]//Proceedings of the Web Conference 2020. 2020: 673-683.

\bibitem{wang2020scalable}
Wang J, Luo M, Suya F, et al. Scalable attack on graph data by injecting vicious nodes[J]. Data Mining and Knowledge Discovery, 2020, 34(5): 1363-1389.

\bibitem{wang2018attack}
Wang X, Cheng M, Eaton J, et al. Attack graph convolutional networks by adding fake nodes[J]. arXiv preprint arXiv:1810.10751, 2018.

\bibitem{entezari2020all}
Entezari N, Al-Sayouri S A, Darvishzadeh A, et al. All you need is low (rank) defending against adversarial attacks on graphs[C]//Proceedings of WSDM'20. 169-177.

\bibitem{wu2019adversarial}
Wu H, Wang C, Tyshetskiy Y, et al. Adversarial examples on graph data: Deep insights into attack and defense[J]. arXiv preprint arXiv:1903.01610, 2019.

\bibitem{jin2020graph}
Jin W, Ma Y, Liu X, et al. Graph structure learning for robust graph neural networks[C]//Proceedings of KDD'20. 66-74.

\bibitem{zhang2020gnnguard}
Zhang X, Zitnik M. Gnnguard: Defending graph neural networks against adversarial attacks[J]. Advances in Neural Information Processing Systems, 2020, 33: 9263-9275.

\bibitem{lim2021new}
Lim D, Li X, Hohne F, et al. New benchmarks for learning on non-homophilous graphs[J]. arXiv preprint arXiv:2104.01404, 2021.

\bibitem{zhu2020beyond}
Zhu J, Yan Y, Zhao L, et al. Beyond homophily in graph neural networks: Current limitations and effective designs[J]. Advances in NeurIPS'20, 33: 7793-7804.

\bibitem{bojchevski2017deep}
Bojchevski A, Günnemann S. Deep gaussian embedding of graphs: Unsupervised inductive learning via ranking[J]. arXiv preprint arXiv:1707.03815, 2017.

\bibitem{mccallum2000automating}
McCallum A K, Nigam K, Rennie J, et al. Automating the construction of internet portals with machine learning[J]. Information Retrieval, 2000, 3(2): 127-163.

\bibitem{giles1998citeseer}
Giles C L, Bollacker K D, Lawrence S. CiteSeer: An automatic citation indexing system[C]//Proceedings of the third ACM conference on Digital libraries. 1998: 89-98.

\bibitem{kipf2016semi}
Kipf T N, Welling M. Semi-supervised classification with graph convolutional networks[J]. arXiv preprint arXiv:1609.02907, 2016.

\bibitem{chen2018fast}
Chen J, Wu Y, Xu X, et al. Fast gradient attack on network embedding[J]. arXiv preprint arXiv:1809.02797, 2018.

\bibitem{chen2020mga}
Chen J, Chen Y, Zheng H, et al. MGA: Momentum gradient attack on network[J]. IEEE Transactions on Computational Social Systems, 2020, 8(1): 99-109.

\bibitem{li2020deeprobust}
Li Y, Jin W, Xu H, et al. Deeprobust: A pytorch library for adversarial attacks and defenses[J]. arXiv preprint arXiv:2005.06149, 2020.

\bibitem{chen2020measuring}
Chen D, Lin Y, Li W, et al. Measuring and relieving the over-smoothing problem for graph neural networks from the topological view[C]//Proceedings of the AAAI'20. 34(04): 3438-3445.

\bibitem{zhao2020data}
Zhao T, Liu Y, Neves L, et al. Data augmentation for graph neural networks[J]. arXiv preprint arXiv:2006.06830, 2020.

\bibitem{jin2021power}
Jin M, Chang H, Zhu W, et al. Power up! robust graph convolutional network via graph powering[C]//35th AAAI Conference on Artificial Intelligence. 2021.

\bibitem{mohri2018foundations}
Mohri M, Rostamizadeh A, Talwalkar A. Foundations of machine learning[M]. MIT press, 2018.
\end{thebibliography}
\end{document}